\documentclass{ecai} 

\usepackage{latexsym}
\usepackage{amssymb}
\usepackage{amsmath}
\usepackage{amsthm}
\usepackage{booktabs}
\usepackage{enumitem}
\usepackage{graphicx}
\usepackage{color}

\usepackage{colortbl}
\usepackage{xcolor} 
\usepackage{multirow}
\usepackage{xr}
\usepackage{hyperref}

\newcommand{\BibTeX}{B\kern-.05em{\sc i\kern-.025em b}\kern-.08em\TeX}

\begin{document}


\colorlet{tablerowcolor}{gray!20}
\newcommand{\rowcol}{\rowcolor{tablerowcolor}} 

\begin{frontmatter}


\paperid{123} 


\title{VectorFit : Adaptive Singular \& Bias Vector Fine-Tuning of Pre-trained Foundation Models}


\author[A]{\fnms{Suhas}~\snm{Hegde}}
\author[A]{\fnms{Shilpy}~\snm{Kaur}}
\author[A]{\fnms{Aruna}~\snm{Tiwari}} 

\address[A]{Indian Institute of Technology Indore}


\begin{abstract}
Popular PEFT methods reduce trainable parameter count for fine-tuning by parameterizing new low-rank or sparse trainable weights in parallel to the frozen pre-trained weights $W$. However, these weights are trained from scratch, and there exists a performance gap between these methods and full fine-tuning, especially in low-budget settings. We introduce VectorFit, a new way of parameterization that efficiently utilizes the existing knowledge embedded in $W$ by adaptively training their singular vectors and biases. We show that utilizing the structural and transformational properties of $W$ in this way can lead to high-rank incremental weight matrices $\Delta W$, comparable to that of full fine-tuning. VectorFit delivers superior results with \textbf{9$\boldsymbol\times$} fewer trainable parameters than the leading PEFT methods. Through comprehensive experiments across 19 datasets covering a wide range of language and vision tasks such as natural language understanding and generation, question answering, image classification, and image generation, we demonstrate that VectorFit surpasses baselines in terms of performance as a function of parameter-efficiency.
\end{abstract}

\end{frontmatter}


\section{Introduction}
\label{sec_intro}
Pre-trained foundation models (PFMs) have set unprecedented standards in language, vision, and audio tasks \cite{llama, ldm, whisper}, showcasing their strong performance across diverse domains. Refining these models through fine-tuning is a powerful approach to enhance their performance across diverse downstream tasks \cite{starcoder, metamath}. This process helps models adhere to given instructions \cite{wizardlm}, adopt preferred behaviors, and discard undesirable ones \cite{direct}. However, adapting these models to downstream tasks through full fine-tuning (Full-FT) is a significant challenge, primarily due to the immense computational and memory overhead. For instance, models like DeBERTa-V3 \cite{debertav3} with 300 million parameters, ViT-22B \cite{vit22b} with 22 billion parameters, and Llama-3 \cite{llama3} with a staggering 405 billion parameters exemplify the scale of modern PFMs. Adapting these models for multiple downstream tasks is resource-intensive and typically requires maintaining separate copies of the full model for each task, leading to a very high memory consumption.

\begin{figure}[t]
\vskip 0.2in
\begin{center}
\centerline{\includegraphics[scale=0.6]{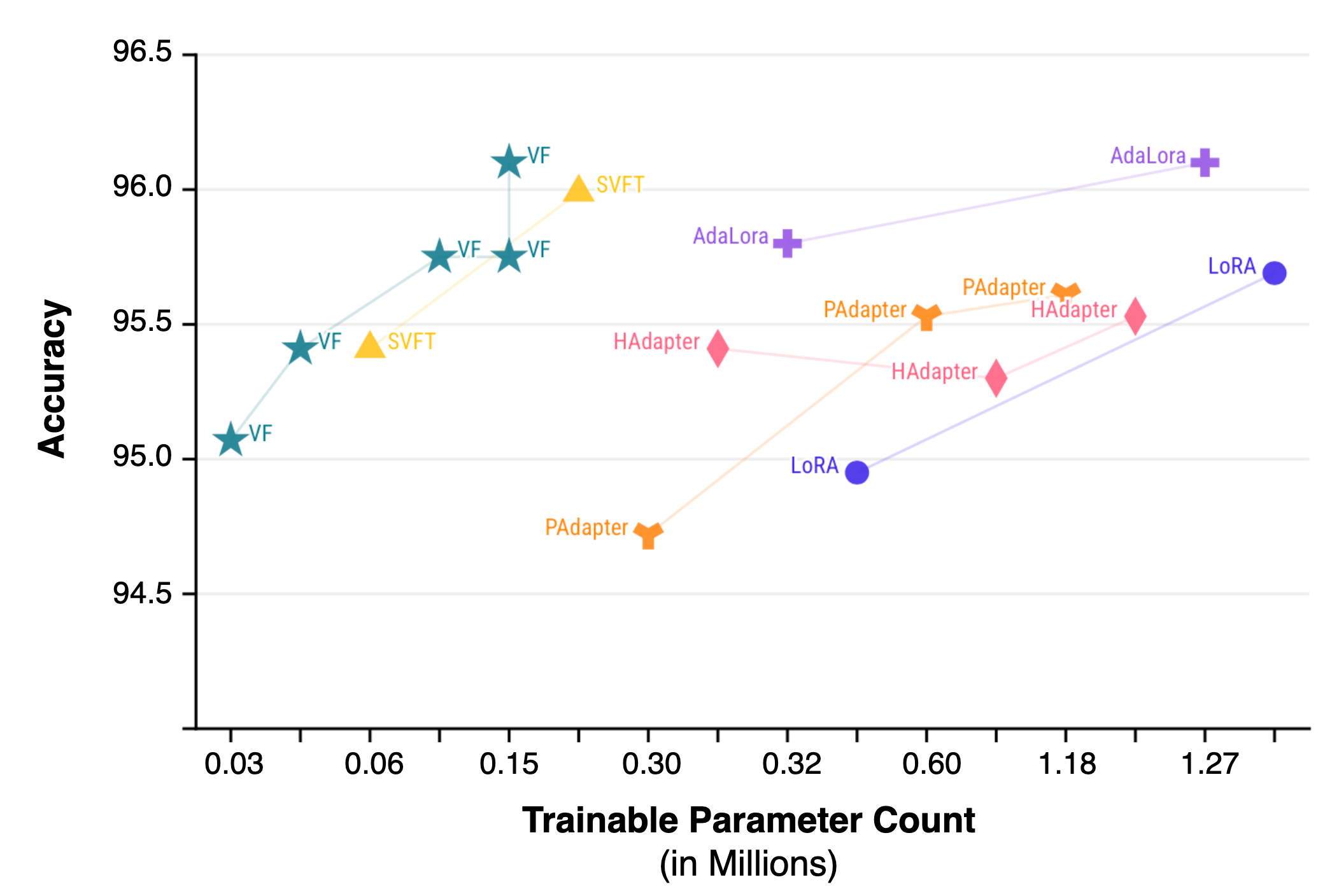}}
\caption{Accuracy vs Trainable parameter count for SST2 dataset. VectorFit (labeled as VF for brevity) outperforms baselines with 85\% less trainable parameters. The graph highlights that VectorFit is a PEFT method in extremely low parameter regime of $<$0.1\% trainable parameters.}
\label{lpr_graph}
\end{center}
\end{figure}

Parameter-Efficient Fine-Tuning (PEFT) mitigates these challenges by introducing a small set of trainable parameters to produce specialized models. For example, PFMs like LLMs are fine-tuned for tasks such as text classification \cite{glue}, question answering \cite{squad}, and text generation \cite{tg_survey} using task-specific datasets. PEFT techniques, such as LoRA \cite{lora} and adapter \cite{padapter}, significantly reduce the number of trainable parameters compared to Full-FT, although this can compromise the performance. More advanced methods like AdaLoRA \cite{adalora} try to increase expressiveness by adaptively choosing trainable parameters, bridging the performance gap. However, majority of successful PEFT methods work by adding a new set of weight matrices with the assumption that incremental weight matrices are low-rank. This may limit their expressiveness and disregards the nuances of weight matrix transformation during Full-FT. Moreover, even state-of-the-art PEFT techniques (e.g., LoRA, Adapter, and AdaLoRA) can still result in a substantial number of trainable parameters, even in their highest parameter-efficient setup (e.g., LoRA with rank 1). Although there are a few recent methods that do not rely on low-rank updates \cite{oft, svft}, all of them, to the best of our knowledge, work based on fine-tuning a newly initialized set of weight matrices. This leads to a high overall parameter count and memory consumption, twice as that of LoRA in SVFT \cite{svft}.

This prompts the question: Can we achieve extreme parameter-efficiency through the efficient use of pre-training knowledge embedded in the weight matrices, without incurring prohibitive parameter and memory costs? As an answer, we introduce VectorFit, which directly leverages the structural and transformational characteristics of the pre-trained weights instead of introducing new weights for fine-tuning, differentiating our method from prior work. Given a pre-trained weight matrix $W_0$, VectorFit applies singular value decomposition (SVD), such that $W_0 = U \Sigma V^T$. The method then selectively adapts the singular vector ($\Sigma$) and the bias ($b$) associated with $W_0$, focusing on those $\Sigma$ and $b$ that exhibit suboptimal training compared to those of other weight matrices. We propose a mechanism called \textit{Adaptive Vector Freezing} to achieve this.

Since the singular vectors represent the stretching of the transformation applied by the weight matrices in their high-dimensional subspace, directly fine-tuning them allows for high expressiveness (Appendix \ref{app_wtransformation}). This is evident from the fact that VectorFit performs high-rank updates comparable to Full-FT (Figure \ref{sin_values}) while using significantly fewer trainable parameters ($\leq 0.1\%$). Additionally, training the bias vectors gives translational degree of freedom, further enhancing the expressiveness during fine-tuning. VectorFit outperforms the baselines in terms of performance relative to parameter efficiency (Figure \ref{lpr_graph}). It also gives a practical memory consumption approximately equivalent to that of LoRA with rank 1 for smaller base models (Figure \ref{memtrace}). The memory consumption of VectorFit lies in between LoRA and SVFT for larger base models (Figure \ref{memuse}).

\begin{figure*}[ht]
\vskip 0.2in
\begin{center}
\centerline{\includegraphics[scale=0.7]{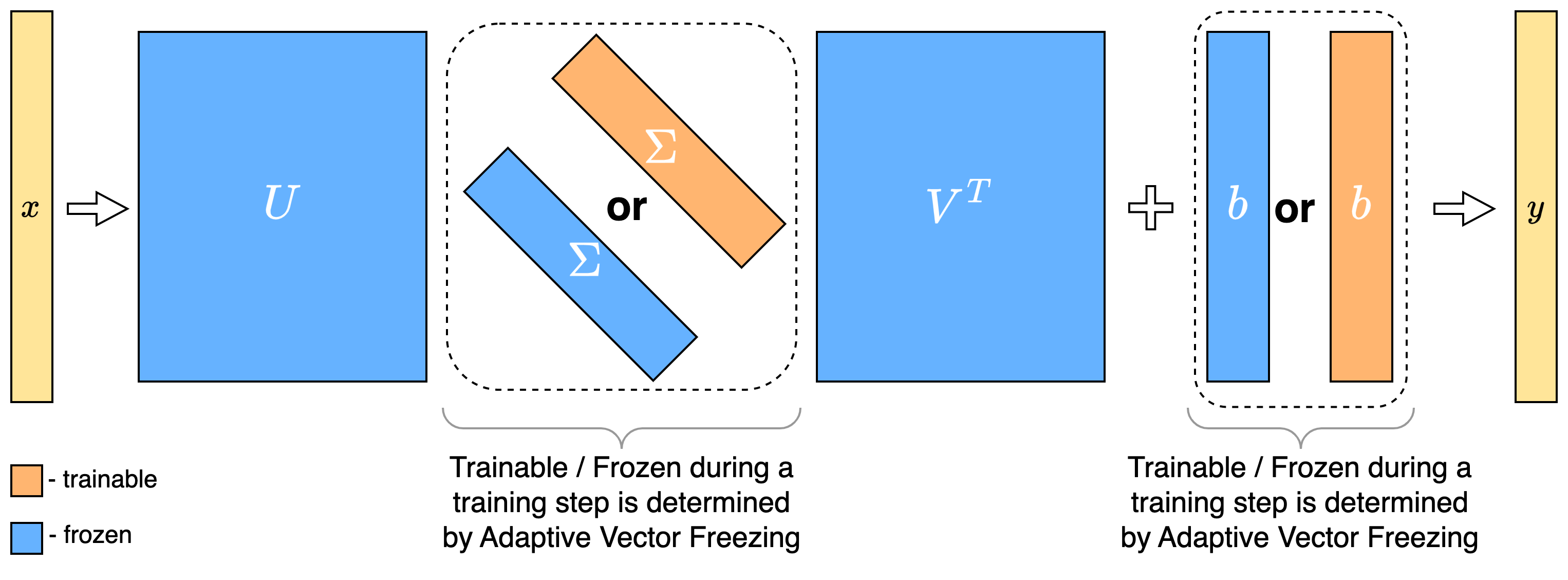}}
\caption{Architecture diagram of VectorFit. The pretrained weight matrix is initially decomposed into $U$, $\Sigma$, and $V$. Subsequently, only $\Sigma$ and bias $b$ are trained with Adaptive Vector Freezing mechanism.}
\label{arch}
\end{center}
\end{figure*} 


\section{Related Work}
\label{sec_rw}

Researchers have explored three primary approaches to reduce the number of parameters required for fine-tuning while preserving or enhancing the performance of PFMs. These approaches can be broadly categorized into Adapter-based methods, LoRA-based methods, and other PEFT methods.

\textbf{Adapter-Based methods.} This research direction emphasizes incorporating small neural networks into PFMs and fine-tuning only these modules for specific tasks, keeping the base model frozen and shared across tasks. This approach introduces a limited number of task-specific parameters, significantly improving the scalability and practicality of large models. For instance, adapter tuning \cite{hadapter, padapter, he2022towards} integrates small neural networks, known as adapters, between the layers of the base model. Other methods, such as prefix tuning \cite{prefix_tuning} and prompt tuning \cite{prompt_tuning}, add trainable prefix tokens to the input or hidden layers of the model. These techniques claim to have demonstrated performance comparable to Full-FT while updating less than 1\% of the model parameters, significantly reducing memory requirements.

\textbf{LoRA-Based methods.} A significant advancement in PEFT is Low-Rank Adaptation (LoRA) \cite{lora}, which preserves the pre-trained model weights and incorporates trainable low-rank matrices within each transformer layer. For a pre-trained weight matrix $W_0 \in \mathbb{R}^{d_r \times d_c}$, LoRA constrains the weight update $\Delta W$ to a low-rank decomposition: $y = W_0 x + \Delta W x = W_0 x + BAx$, where $B \in \mathbb{R}^{d_r \times r}$, $A \in \mathbb{R}^{r \times d_c}$ and rank $r<<min(d_r,d_c)$. Only $A$ and $B$ are trainable parameters.

Several studies have introduced variations of the LoRA algorithm, focusing on reducing the number of trainable parameters \cite{lorafa, vera, sparse}, improving the flexibility of low-rank structures \cite{nola, deltalora, svfit}, enabling adaptive parameter allocation \cite{increlora}, 
and integrating LoRA with techniques like quantization \cite{qlora, qalora} and pruning \cite{loraprune}.

A significant enhancement over LoRA is AdaLoRA \cite{adalora}, which addresses LoRA's limitation of evenly distributing trainable parameters across weight matrices, ignoring their varying importance. In AdaLoRA, the incremental updates are parameterized as singular value decomposed matrices $P \Lambda Q$, where $P \in \mathbb{R}^{d_r \times r}$ and $Q \in \mathbb{R}^{r \times d_c}$ are the left and right singular matrices, $\Lambda \in \mathbb{R}^{r \times 1}$ is the singular vector. The orthogonality of $P$ and $Q$ is maintained using the regularizer $R(P,Q) = \|P^\top P - I\|^2_F + \|Q Q^\top - I\|^2_F$. The rank of low-rank updates is dynamically adjusted using an importance metric derived from $\Lambda$. By pruning less significant singular values while allowing for recovery, AdaLoRA claims to have improved performance with a similar parameter budget as LoRA. Pissa \cite{pissa} is another method that works similar to LoRA while the initialization of the low-rank trainable weights is done using the principal singular values and vectors. 

\textbf{Other PEFT methods.} Orthogonal Fine-Tuning (OFT) \cite{oft} introduces an orthogonal projection approach using orthogonal regularization. It focuses on optimizing parameters while preserving the orthogonality of weight updates, ensuring minimal interference with pre-trained knowledge. However, it still demands a significant number of trainable parameters because of the high dimensionality of the matrices. Butterfly Orthogonal Fine-Tuning (BOFT) \cite{boft} builds upon OFT by introducing Butterfly factorization and claims to improve parameter efficiency, and fine-tuning flexibility. Singular Vectors guided Fine-Tuning (SVFT) \cite{svft} leverages the singular value decomposition of pre-trained weight matrices to parameterize weight updates as $y = W_0 x + \Delta Wx = U(\Sigma + M)V^T x$, where $M$ is a sparse trainable matrix with pre-determined and fixed sparsity pattern. As $M$ is not restricted to be low-rank, SVFT claims to achieve high-rank gradient updates. Nonetheless, SVFT uses four matrices, \( U, \Sigma, V \), and \( M \), for every pre-trained weight matrix. Also, their dimensions are comparable to those of the pre-trained weight matrix. This leads to a high parameter and memory cost. 


\section{VectorFit}
\label{sec_vf}

In this section, we describe VectorFit and its components in detail. VectorFit comprises two key components: (1) Vector Fine-Tuning, based on SVD. (2) \textit{Adaptive Vector Freezing}, a mechanism to avoid co-adaptation and to improve the performance.

\subsection{Vector Fine-Tuning}
\label{sec_vft}
VectorFit initially performs SVD on the pre-trained weight matrix $W_0 \in \mathbb{R}^{d_r \times d_c}$, such that, $W_0 = U \Sigma V^T$. $W_0$ can be the weight matrix of any of the modules in self-attention ($q, k, v, o$) or multilayer perceptron ($f_1, f_2$) of a transformer block. Then we potentially fine-tune only the singular vector $\Sigma$ and the pre-trained bias vector $b_0$ corresponding to $W_0$, subject to \textit{Adaptive Vector Freezing}, as shown in Figure \ref{arch}. Formally, this can be denoted as follows:

\begin{equation}
\label{forward}
y = \left( U \textcolor[rgb]{1.00,0.00,0.00}{\Sigma} V^T \right) x + \textcolor[rgb]{1.00,0.00,0.00}{b_0} 
\end{equation}

where $x \in \mathbb{R}^{d_{in} \times d_r}$ is the input hidden state. $U \in \mathbb{R}^{d_r \times d_r}$ is the left singular matrix and $V \in \mathbb{R}^{d_c \times d_r}$ is the right singular matrix of $W_0$, consisting of orthonormal column vectors. $V^T$ is the transpose of V. $\Sigma \in \mathbb{R}^{d_r \times 1}$ and $b_0 \in \mathbb{R}^{d_c}$ are the potentially trainable singular vector and bias vector of $W_0$, respectively. $y \in \mathbb{R}^{d_{out} \times d_c}$ is the output hidden state. Note that $\Sigma$ in standard SVD is a diagonal matrix and we store it as a vector for memory efficiency. The weight updates of VectorFit are parameterized as:

\begin{equation}
\label{sin_vec_training}
W = W_0 + \Delta W = U(\Sigma + \Delta\Sigma)V^T
\end{equation}
\begin{equation}
\label{bias_training}
b = b_0 + \Delta b
\end{equation}

where $\Delta W$, $\Delta \Sigma$, and $\Delta b$ are the incremental matrix/vectors of $W_0$, $\Sigma$, and $b_0$, respectively. 

Singular values of a weight matrix quantify the scaling factors for the transformation along its orthogonal directions, an important aspect of the weight matrix. Directly fine-tuning them as described above results in an overall high-rank incremental matrix whose rank is comparable to that of full fine-tuning of the weight matrix. This is analyzed in detail in Section \ref{sec_dis_ra}.

To avoid performing the expensive calculation of SVD for every single pre-trained weight matrix during each training step, we perform SVD in the beginning of the fine-tuning and replace the original weight matrices of the model with their decomposed version. This takes a few seconds of extra time in the beginning of the fine-tuning, which is negligible. Although this approach increases the total parameter count---for instance, VectorFit with DeBERTaV3-base has 18\% more parameters than LoRA (r = 1) with DeBERTaV3-base---its practical training memory consumption remains similar to LoRA (r = 1). More details on this is given in Appendix \ref{app_A}.

\subsection{Adaptive Vector Freezing}
\label{sec_avf}
As we train only a small number of parameters (a few tens of singular and bias vectors), it is crucial to ensure balanced training across all trainable vectors. This prevents some vectors from being over-trained while others remain under-trained, a phenomenon known as co-adaptation.

To address this, we propose \textit{Adaptive Vector Freezing (AVF)}, a mechanism that periodically freezes (disables the gradients of) the top-$k$ trainable vectors that have undergone extensive training. This allows the remaining under-trained vectors to receive adequate updates. The extent of training of each vector is quantified in the form of training strength. Consider the set of all trainable vectors, $V = \{\Sigma_{l,m},b_{l,m} : l = layer, m = module(q,k,v,o,f_1,f_2)\}$. We define the training strength $S_{v}(t)$ of a vector $v \in V$ at training step $t$ as L1 norm between $v_0$ and $v_t$, formulated as follows:

\begin{equation}
\label{training_strength}
S_{v}(t) = \frac{1}{dim(v)}{|| v_0 - v_t ||}_1
\end{equation}

\begin{table*}[ht]
\caption{DeBERTaV3-base fine-tuned using various PEFT methods is evaluated on the GLUE benchmark. For performance metrics, we report matched accuracy for MNLI, Matthew’s correlation for COLA, Pearson correlation for STS-B, and accuracy for the other tasks, where higher values indicate better performance across all metrics. \# Params is the number of trainable parameters.}
\vspace{0.5cm}
\label{glue-table}
\centering
\begin{tabular}{c | c | c  c  c  c  c  c  c  c}
\toprule
\textbf{Method}   & \textbf{\# Params} & \textbf{MNLI}        & \textbf{SST2}  & \textbf{COLA}  & \textbf{QQP}         & \textbf{QNLI}  & \textbf{RTE}   & \textbf{MRPC}  & \textbf{STSB}  \\ \midrule
Full FT   & 184M   & 89.90/90.12 & 95.63 & 69.19 & 92.40/89.80 & 94.03 & 83.75 & 89.46 & 91.60 \\ \midrule
HAdapter  & 1.22M  & 90.13/90.17 & 95.53 & 68.64 & 91.91/89.27 & 94.11 & 84.48 & 89.95 & 91.48 \\
PAdapter  & 1.18M  & 90.33/90.39 & 95.61 & 68.77 & 92.04/89.40 & 94.29 & 85.20 & 89.46 & 91.54 \\
LoRA(r=8) & 1.33M  & 90.65/90.69 & 94.95 & 69.82 & 91.99/89.38 & 93.87 & 85.20 & 89.95 & 91.60 \\
AdaLora   & 1.27M  & \textbf{90.76/90.79} & \textbf{96.10} & \textbf{71.45} & \textbf{92.23/89.74} & \textbf{94.55} & \textbf{88.09} & \textbf{90.69} & \textbf{91.84} \\ \midrule
HAdapter  & 0.61M  & 90.12/90.23 & 95.30 & 67.87 & 91.65/88.95 & 93.76 & 85.56 & 89.22 & 91.30 \\
PAdapter  & 0.60M  & 90.15/90.28 & 95.53 & 69.48 & 91.62/88.86 & 93.98 & 84.12 & 89.22 & 91.52 \\
HAdapter  & 0.31M  & 90.10/90.02 & 95.41 & 67.65 & 91.54/88.81 & 93.52 & 83.39 & 89.25 & 91.31 \\
PAdapter  & 0.30M  & 89.89/90.06 & 94.72 & 69.06 & 91.40/88.62 & 93.87 & 84.48 & 89.71 & 91.38 \\
LoRA(r=2) & 0.33M  & 90.30/90.38 & 94.95 & 68.71 & 91.61/88.91 & 94.03 & 85.56 & 89.71 & 91.68 \\
AdaLora   & 0.32M  & \textbf{90.66/90.70} & 95.80 & 70.04 & \textbf{91.78/89.16} & \textbf{94.49} & 87.36 & 90.44 & 91.63 \\ 
Pissa & 0.33M  & 89.99/90.13 & 94.86 & 68.79 & 91.57/88.73 & 93.92 & 85.09 & 89.84 & 91.69 \\
SVFT    & 0.28M   & 89.90/89.97 & 95.99 & \textbf{72.61} & 91.50/88.98 & 93.90 & \textbf{88.09} & 88.99 & 91.73 \\ 
\rowcol VectorFit & 0.15M & \underline{90.12/89.89} & \underline{\textbf{96.10}} & \underline{70.94} & \underline{91.51/88.70} & \underline{94.05} & \underline{84.12} & \underline{\textbf{92.16}} & \underline{\textbf{91.76}} \\ 
\bottomrule
\end{tabular}
\end{table*}

where $v_0$ is the value of $v$ before fine-tuning and $v_t$ is the value of $v$ during the training step $t$. $dim(v)$ is the dimension of $v$. To find the top-$k$ vectors, we perform exponential moving average of $S_{v}(t)$ as given in Eq. \ref{ema}.

\begin{equation}
\label{ema}
S'_{v}(t) = \beta S'_{v}(t - t_f) + (1 - \beta) S_{v}(t)
\end{equation}

where $\beta = 0.99$, is a constant. We define the training step at which AVF is applied as the AVF step. Given the frequency of AVF steps $t_f$, the first AVF step $t_i$, the number of vectors $k$ to freeze per AVF step, and the total number of AVF steps $n_f$, the top-$k$ vectors with the highest $S'_v(t)$ values are frozen at each AVF step. Note that the trainability of vectors do not change in between AVF steps. However, a vector frozen during one AVF step may become trainable in a subsequent AVF step, ensuring that all vectors are adequately trained over time. More details on these hyperparameters are given in Appendix \ref{app_impl}.

In Section \ref{sec_dis_avf}, we theoretically and experimentally show that this mechanism leads to similar effect as that of dropout. It is important to note that using the standard dropout algorithm for singular vectors results in a significant performance drop, even with a very low dropout probability. This signifies that certain singular values, and their corresponding left and right singular directions are extremely important that they cannot be dropped. Therefore, the AVF mechanism is crucial for maintaining effective training.

\section{Experiments}
\label{sec_exp}
\textbf{Implementation details.} All algorithms are implemented using PyTorch \cite{pytorch}. Our implementation builds upon the publicly available Huggingface Transformers codebase \cite{hft}. Appendix \ref{app_impl} contains the full details of our experimental setup and hyperparameter configurations.

\subsection{Tasks and Datasets}
\label{sec_td}
We evaluate our method on the following tasks and datasets: 
\begin{enumerate}
    \item Natural Language Understanding (NLU): Experiments are conducted on the GLUE benchmark \cite{glue}, which includes single-sentence classification, similarity/paraphrase, and natural language inference tasks.
    \item Question Answering (QA): Performance is tested on SQuAD v1.1 and SQuAD v2.0 \cite{squad}, treating QA as a sequence labeling problem to predict the start and end token probabilities for answer spans.
    \item Natural Language Generation (NLG): Evaluation is performed on the XSum \cite{xsum} and CNN/DailyMail \cite{cnn} datasets for text summarization task. GSM8K \cite{gsm8k} and Math \cite{math} datasets are used to evaluate mathematical problem-solving capabilities.
    \item Image Classification: Our method is assessed on image classification tasks using CIFAR10 \cite{cifar10}, GTSRB \cite{gtsrb}, MNIST \cite{mnist}, and RESISC45 \cite{resisc45} datasets.
    \item Image Generation: The results on subject-driven image generation is evaluated using the Dreambooth dataset \cite{dreambooth}.
\end{enumerate}

\subsection{Base Models}
\label{sec_bm}
We use six distinct base model types that are representative of a wide range of PFMs for the evaluation of our algorithm: 
\begin{enumerate}
    \item DeBERTaV3-base \cite{debertav3}, a transformer encoder-only language model, applied to NLU and QA tasks.
    \item BART-large \cite{bart}, a transformer encoder-decoder model, used for NLG tasks. 
    \item Gemma-7B \cite{gemma}, a transformer decoder-only model, used for NLG tasks. 
    \item Llama-3-8B \cite{llama3}, a transformer decoder-only model, used for NLG tasks. 
    \item ViT-base \cite{vit}, a vision transformer model, applied to image classification tasks, pre-trained on Imagenet-1K.
    \item Stable Diffusion v1.4 \cite{ldm}, a latent diffusion model based on UNet architecture, used for text-to-image generation.
\end{enumerate}
\textbf{Baselines.} We compare our approach against Full-FT, which updates all parameters across all layers. Additionally, we evaluate it against state-of-the-art methods from each of the three categories mentioned in Section \ref{sec_rw}. These include LoRA \cite{lora}, AdaLoRA \cite{adalora}, PAdaptor \cite{padapter}, HAdaptor \cite{hadapter}, Pissa \cite{pissa}, and SVFT \cite{svft}. 

\section{Results}
\label{sec_res}
In the tables, the highest accuracy within each trainable parameter count regime is highlighted in \textbf{bold}, and the overall parameter efficiency (\% accuracy / \% trainable parameters) is reported with \underline{underline}.

\subsection{Natural Language Understanding}
\label{sec_res_nlu}

Table \ref{glue-table} presents the results on the GLUE benchmark, where VectorFit outperforms Full-FT by an average of 0.6\% while requiring over 1200$\times$ fewer trainable parameters. Its performance is comparable to baselines like LoRA (r = 8), which uses 9$\times$ more trainable parameters. VectorFit outperforms SVFT by upto 3.2\% with 2$\times$ less parameters. Notably, VectorFit achieves the highest parameter efficiency across all datasets in the GLUE benchmark, establishing it as the most optimal PEFT method for natural language understanding tasks.

\subsection{Question Answering}
\label{sec_res_qa}

We evaluate the performance of our method on the SQuAD v1.1 and the more challenging SQuAD v2.0 datasets, using exact match (EM) and F1 scores as metrics. The results, summarized in Table \ref{qa-table}, demonstrate that VectorFit outperforms the baselines on SQuAD v1.1 giving 0.9\% better F1 score on an average. It achieves superior results compared to Full-FT with 1250$\times$ fewer parameters. On SQuAD v2.0, VectorFit delivers performance comparable to Full-FT and the best-performing baselines, highlighting its efficiency and effectiveness.

\begin{table}[!h]
\vspace{-0.1cm}
\caption{Performance results for DeBERTaV3-base fine-tuned on SQuAD v1.1 and SQuAD v2.0 are presented. \# Params indicates the percentage of trainable parameters. The metrics reported are Exact Match and F1 scores (EM/F1).}
\label{qa-table}
\vspace{0.5cm}
\centering
\resizebox{\columnwidth}{!}{
\begin{tabular}{c | c | c}
\toprule
\textbf{Model} & \textbf{Squad v1.1 (EM/F1)} & \textbf{Squad v2.0 (EM/F1)} \\ \midrule
Full FT        & 86.0 / 92.7                 & 85.4 / 88.4                   \\ \midrule
\# Params      & 0.08\%                      & 0.08\%                      \\ \midrule
HAdapter       & 84.4 / 91.5                 & 83.4 / 86.6                   \\
PAdaptor       & 84.4 / 91.7                 & 84.2 / 87.2                   \\
LoRA           & 86.4 / 92.8                 & 84.6 / 87.5                   \\
AdaLora        & 86.8 / 93.0                 & \textbf{84.7 / 87.6}          \\
Pissa           & 85.9 / 92.3                 & 84.2 / 87.3                   \\
SVFT           & 86.3 / 92.5                 & 84.3 / 87.3                   \\
\rowcol VectorFit & \textbf{87.0 / 93.2}     & 84.4 / \textbf{87.6}          \\ \bottomrule
\end{tabular}
}
\end{table}

\begin{table*}[!h]
\caption{Performance results for BART-large fine-tuned on the XSum and CNN/DailyMail datasets are shown. The \# Params column represents the percentage of trainable parameters. The reported metrics are ROUGE-1, ROUGE-2, and ROUGE-L (R-1/2/L).}
\label{nlg-table}
\vspace{0.5cm}
\centering
\begin{tabular}{c | c | c | c}
\toprule
\textbf{Method}   &\textbf{\# Params} & \textbf{Xsum} & \textbf{CNN/Dailymail} \\ \midrule
Full FT           & 100\%    & 45.49 / 22.33 / 37.26 & 44.16 / 21.28 / 40.90 \\ \midrule
PAdapter          & 0.16\%   & 40.21 / 18.92 / 32.34 & 41.96 / 19.47 / 38.10 \\
LoRA              & 0.16\%   & 42.81 / 19.68 / 34.73 & 43.68 / 20.63 / 40.71 \\ 
AdaLoRA           & 0.16\%   & 43.29 / 19.95 / 35.04 & 43.94 / 20.83 / 40.96 \\
Pissa             & 0.16\%   & 42.67 / 19.51 / 34.16 & 42.97 / 20.04 / 40.11 \\
SVFT              & 0.15\%   & \textbf{43.30} / 19.82 / 35.13 & 43.87 / 20.72 / 40.80 \\
\rowcol VectorFit & 0.12\%   & \underline{43.28 / \textbf{20.71} / \textbf{35.42}} & \underline{\textbf{44.01} / \textbf{21.60} / \textbf{40.98}} \\ \bottomrule
\end{tabular}
\end{table*}

\begin{table*}[!h]
\caption{Comparison of various methods for Gemma-7B and Llama-3-8B models, fine-tuned for mathematical reasoning. The table includes number of parameters and accuracy on GSM-8k and MATH benchmarks.}
\vspace{0.5cm}
\centering
\begin{tabular}{c|ccc|ccc}
\toprule
\multirow{2}{*}{\textbf{Method}} & \multicolumn{3}{c|}{\textbf{Gemma-7B}}               & \multicolumn{3}{c}{\textbf{Llama-3-8B}}              \\ \cmidrule(l){2-7} 
                                 & \textbf{\# Params} & \textbf{GSM-8k} & \textbf{MATH} & \textbf{\# Params} & \textbf{GSM-8k} & \textbf{MATH} \\ \midrule
Full FT       & 8.5B  & 74.67          & 25.70          & 8.0B  & 64.13          & 16.24          \\ \midrule
LoRA (r=1)    & 0.82M & 72.4           & 26.28          & 1.77M & 68.84          & 20.94          \\
AdaLoRA (r=1) & 0.81M & 72.5           & 26.41          & 1.77M & 68.73          & \textbf{21.09}          \\
Pissa          & 0.82M & 72.3           & 26.31          & 1.77M & 67.92          & 21.01 \\
SVFT          & 0.43M & 73.50          & 27.30          & 0.48M & 69.22          & 20.44          \\
\rowcol VectorFit     & 0.43M & \underline{\textbf{73.94}}          & \underline{\textbf{27.41}} & 0.48M & \underline{\textbf{70.83}} & \underline{21.02}          \\ \bottomrule
\end{tabular}
\label{tab:comp}
\end{table*}

\begin{table*}[!h]
\caption{The performance results for ViT-base fine-tuned on the CIFAR10, GTSRB, MNIST, and RESISC45 datasets are presented. The \# Params column indicates the proportion of trainable parameters, and the corresponding image classification accuracies are reported. Section \ref{sec_dis_abl} contains more details about the variations of VectorFit.}
\label{ic-table}
\vspace{0.5cm}
\centering
\begin{tabular}{c|c|cccc}
\toprule
\textbf{Method} & \textbf{\# Params} & \textbf{CIFAR10} & \textbf{GTSRB} & \textbf{MNIST} & \textbf{RESISC45} \\ \midrule
Full-FT        & 100\%  & 98.5 & 99.2 & 99.8 & 95.7 \\ \midrule
LoRA           & 0.3\%  & 98.4 & 99.2 & \textbf{99.7} & \textbf{95.8} \\
AdaLoRA        & 0.3\%  & 98.6 & 99.3 & 99.6 & \textbf{95.8} \\
SVFT           & 0.3\%  & 98.7 & 99.5 & 99.6 & 95.0 \\
\rowcol VectorFit ($\Sigma$) & 0.06\% & \underline{98.6} & \underline{98.0} & \underline{98.3} & \underline{92.2} \\
\rowcol VectorFit (\textit{no avf})  & 0.1\%  & 99.0 & 99.6 & 99.0 & 94.4 \\
\rowcol VectorFit      & 0.1\%  & \textbf{99.1} & \textbf{99.8} & 99.4 & 95.1 \\ \bottomrule
\end{tabular}
\end{table*}

\subsection{Natural Language Generation}
\label{sec_res_nlg}

We evaluate VectorFit on the XSum and CNN/DailyMail datasets using the ROUGE (1/2/L) metrics in Table \ref{nlg-table}. Despite a 33.3\% higher relative parameter efficiency than the baselines, VectorFit consistently outperforms them on both datasets. Notably, VectorFit recovers 95\% of Full-FT Rouge-L score with only 0.12\% trainable parameters, compared to the baselines that recover 86\% accuracy with 0.16\% trainable parameters on the Xsum dataset. VectorFit's performance surpasses Full-FT on CNN/Dailymail dataset, demonstrating superior efficiency and performance on complex tasks. Additionally, we evaluate VectorFit for mathematical reasoning with Gemma-7B and Llama-3-8B, as shown in Table \ref{tab:comp}, highlighting that our method scales well to larger base model sizes.

\subsection{Image Classification}
\label{sec_res_ic}

Table \ref{ic-table} showcases the results on image classification tasks. Our method surpasses Full-FT performance with only 0.1\% trainable parameters. VectorFit achieves comparable results to the baselines while maintaining 80\% higher relative parameter efficiency. Notably, VectorFit ($\Sigma$) demonstrates the highest parameter efficiency, with an average accuracy reduction of just 1.5\% compared to Full-FT.

\begin{table}[!h]
\caption{Quantitative evaluation of Stable Diffusion v1.4 fine-tuned with PEFT methods using Dreambooth approach for Subject-driven Image generation. We evaluate subject fidelity using DINO and CLIP-I, and prompt fidelity using CLIP-T. Higher values indicate better performance across all metrics. \# Params indicates the percentage of trainable parameters.}
\label{ig-table}
\vspace{0.5cm}
\centering
\begin{tabular}{c|c|ccc}
\toprule
\textbf{Method} & \textbf{\# Params} & \textbf{DINO} & \textbf{CLIP-I} & \textbf{CLIP-T} \\ \midrule
Full-FT         & 100\%              & 0.651         & 0.817           & 0.293           \\
LoRA         & 0.04\%              & 0.636         & 0.789           & 0.286           \\
\rowcol VectorFit       & 0.04\%             & 0.642         & 0.796           & 0.289           \\ \bottomrule
\end{tabular}
\end{table}

\begin{figure*}[!h]
\vskip 0.2in
\begin{center}
\centerline{\includegraphics[scale=0.5]{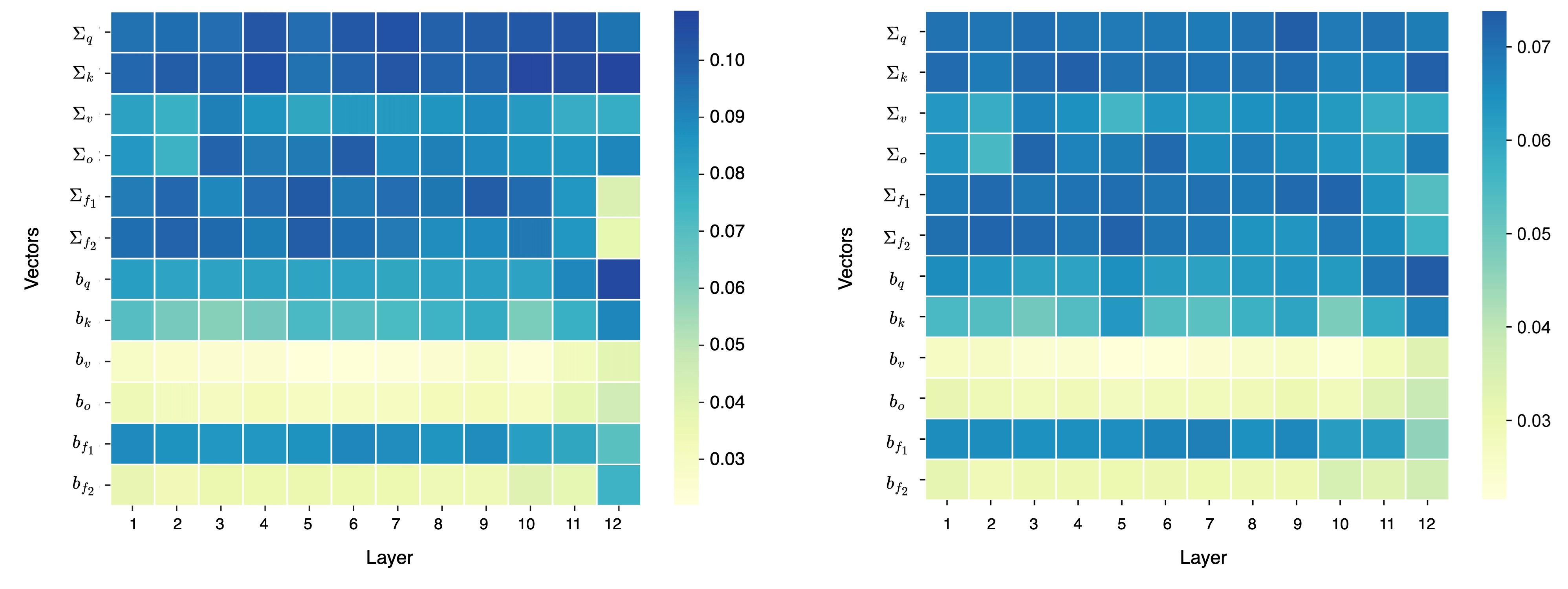}}
\caption{The training strength $S_v$ of each trainable vector after fine-tuning of DeBERTaV3-base on the COLA dataset is shown for VectorFit without AVF (left) and with AVF (right). The x-axis represents the layer index, while the y-axis corresponds to different types of trainable vectors. The heatmaps show the regularization effect (overall lower $S_v$ values) and the balanced training achieved with AVF.}
\label{hm_noavf}
\end{center}
\end{figure*}

\subsection{Image Generation}
\label{sec_res_ig}
Table \ref{ig-table} summarizes the results of personalized image generation using Dreambooth-style fine-tuning \cite{dreambooth}. The evaluation is conducted using three metrics: DINO, CLIP-I, and CLIP-T, as outlined in \cite{dreambooth}. Our method recovers an average DINO, CLIP-I, and CLIP-T score of 98.2\% of Full-FT accuracy as opposed to 97.3\% achieved by LoRA with 0.04\% of trainable parameters. Figure \ref{db_qual} provides a visual comparison between Full-FT and VectorFit.

\section{Discussion}
\label{sec_dis}
\subsection{Effect of Adaptive Vector Freezing}
\label{sec_dis_avf}

\textit{\textbf{Proposition 1.} AVF has an effect comparable to that of dropout.} 

\textit{\textbf{Proof.}} Let $n_u$ represent the total number of training/gradient update steps. The gradient of a vector $v$ with respect to loss $L$ is $\nabla_v L$. The expected gradient for the vector $v$ per step without AVF is expressed as:
\begin{equation}
\label{orig_expec}
\mathop{\mathbb{E}} \left[ \nabla_v L \right] = \frac{1}{n_u} \sum_{i = 1}^{n_u}\left( \nabla_v L \right)_{[i]}
\end{equation}

With $n_f$ AVF steps, the training process can be divided into $n_f + 1$ gradient update intervals. Let $p_j$ denote the probability that a vector $v$ is frozen during the interval $j$. Let $i_s$ and $i_e$ be the temporary variables that denote the first and last gradient update step within each interval respectively. The expected gradient update per step for $v$ under AVF can be expressed as:

\begin{equation}
\label{freeze_expec}
{\mathop{\mathbb{E}}}_{f} \left[ \nabla_v L \right] = \frac{1}{n_u} \sum_{j = 1}^{n_f + 1}\sum_{i = i_s}^{i_e}(1 - p_j)\left( \nabla_v L \right)_{[i]} \\
\end{equation}
\begin{equation}
\label{freeze_expec}
 = \frac{1}{n_u} \left[ \sum_{i = 1}^{n_u}\left( \nabla_v L \right)_{[i]} - \sum_{j = 1}^{n_f + 1}\sum_{i = i_s}^{i_e}(p_j)\left( \nabla_v L \right)_{[i]} \right]
\end{equation}
\begin{equation}
\label{freeze_expec}
\mathop{\mathbb{E}_f} \left[ \nabla_v L \right] = \mathop{\mathbb{E}} \left[ \nabla_v L \right] - \frac{1}{n_u} \sum_{j = 1}^{n_f + 1}\sum_{i = i_s}^{i_e}(p_j)\left( \nabla_v L \right)_{[i]}
\end{equation}

The second term of Eq. \ref{freeze_expec} captures the regularization effect of AVF. A similar analysis can be applied to dropout, demonstrating that AVF effectively minimizes co-adaptation. This effect is empirically validated in Figure \ref{hm_noavf}.

\subsection{Rank Analysis}
\label{sec_dis_ra}

\textit{\textbf{Proposition 2:} If the rank of a matrix is high, it contains more information compared to its low-rank counterpart.}

\textbf{\textit{Proof.}}
Let $A$ be a matrix of dimensions $m \times n$ with rank $p$. Suppose $A$ is decomposed using Singular Value Decomposition (SVD) as:
\[
A = U \Sigma V^T,
\]
where $U$ and $V$ are orthonormal matrices, and $\Sigma$ is a diagonal matrix containing singular values.

Now, consider a lower-rank approximation of $A$, denoted as $A_k$, with rank $k$ such that $k < p$. The truncated SVD of $A_k$ is given by:
\[
A_k = U_k \Sigma_k V_k^T,
\]
where $U_k$, $V_k$, and $\Sigma_k$ correspond to the top $k$ singular components of $A$.

The difference in Frobenius norm between $A$ and $A_k$ is given by:
\[
\| A - A_k \|_F = \sqrt{\sum_{i=k+1}^{p} \sigma_i^2},
\]
where $\sigma_i$ are the singular values of $A$.

This difference represents the information loss due to low-rank approximation. Since the Frobenius norm quantifies the variance within the matrix, reducing the rank leads to a loss in variance information. Therefore, a higher-rank matrix retains more information than its lower-rank approximation.

\textit{\textbf{Proposition 3:} Singular vector updates lead to high-rank incremental matrices.}

\textbf{\textit{Proof.}}
Consider the incremental weight matrix $\Delta^*$:
\[
\Delta^* = W_{\text{init}} - W_{\text{final}}
\]
Expressing in terms of singular value decomposition:
\[
\Delta^* = U \Sigma_{\text{init}} V^T - U \Sigma_{\text{final}} V^T
\]

\[
\Delta^* =  U \Delta \Sigma V^T
\]
Since $U$ and $V$ are orthogonal matrices, the rank of $\Delta^*$ is upper bounded by the rank of $\Delta \Sigma$. This ensures that $\Delta^*$ achieves a full-rank incremental update when all singular values are updated, meaning no singular values are truncated.

Furthermore, by the Eckart–Young–Mirsky theorem \cite{eymtheorem}, the best rank-$k$ approximation of a matrix $A$ is obtained by retaining the top $k$ singular values. Conversely, if no singular values are truncated (i.e., all singular values are updated or remain nonzero), the approximation $\Delta^*$ preserves the full rank:
\[
r = \min(d_r, d_c).
\]
where $d_r$ and $d_c$ are the dimensions of weight matrix. This confirms that singular vector updates lead to high-rank incremental matrices. We experimentally verify this by plotting the singular values of $\Delta^*$ from randomly selected layers as shown in Figure \ref{sin_values} of Appendix \ref{app_mexp_rank}.

\begin{table*}[!h]
\caption{Ablation study to analyze the efficacy of AVF. The table presents results for L1 regularization, Random Vector Freezing, and AVF.}
\vspace{0.5cm}
\centering
\begin{tabular}{@{}c|ccccc|cc@{}}
\toprule
\textbf{Method} & \textbf{SST-2} & \textbf{CoLA} & \textbf{RTE} & \textbf{MRPC} & \textbf{STSB} & \textbf{SQuAD v1.1} & \textbf{SQuAD v2.0} \\ \midrule
L1 regularization                & 90.02 & 64.74 & 69.38 & 74.77 & 84.53 & 71.3 / 74.5 & 60.9 / 63.1 \\
Random vector freezing & 94.16 & 70.13 & 82.82 & 91.26 & 90.17 & 85.1 / 91.7 & 82.8 / 85.9 \\
Adaptive vector freezing (AVF)              & 96.10 & 70.94 & 84.12 & 92.16 & 91.76 & 87.0 / 93.2 & 84.4 / 87.6 \\ \bottomrule
\end{tabular}
\label{l1_avf}
\end{table*}

\subsection{Ablations on Choice of Vectors and AVF}
\label{sec_dis_abl}
This section presents the experiments that reveal the contribution of different vectors and the AVF mechanism to the performance of VectorFit. To this end, we explore 5 variants of VectorFit.

\textit{VectorFit ($\Sigma_a$)}:
The singular vectors corresponding to \{$q,k,v,o$\} are trained.

\textit{VectorFit ($\Sigma$)}:
The singular vectors corresponding to \{$q,k,v,o,f_1,f_2$\} are trained.

\textit{VectorFit ($\Sigma_a + b$)}:
The singular vectors corresponding to \{$q,k,v,o$\} and all the bias vectors are trained.

\textit{VectorFit (\textit{no avf})}:
The singular vectors corresponding to \{$q,k,v,o,f_1,f_2$\} and all the bias vectors are trained.

\textit{VectorFit}:
The singular vectors corresponding to \{$q,k,v,o,f_1,f_2$\} and all the bias vectors are trained along with AVF.

\begin{figure}[ht]
\centerline{\includegraphics[width=\columnwidth]{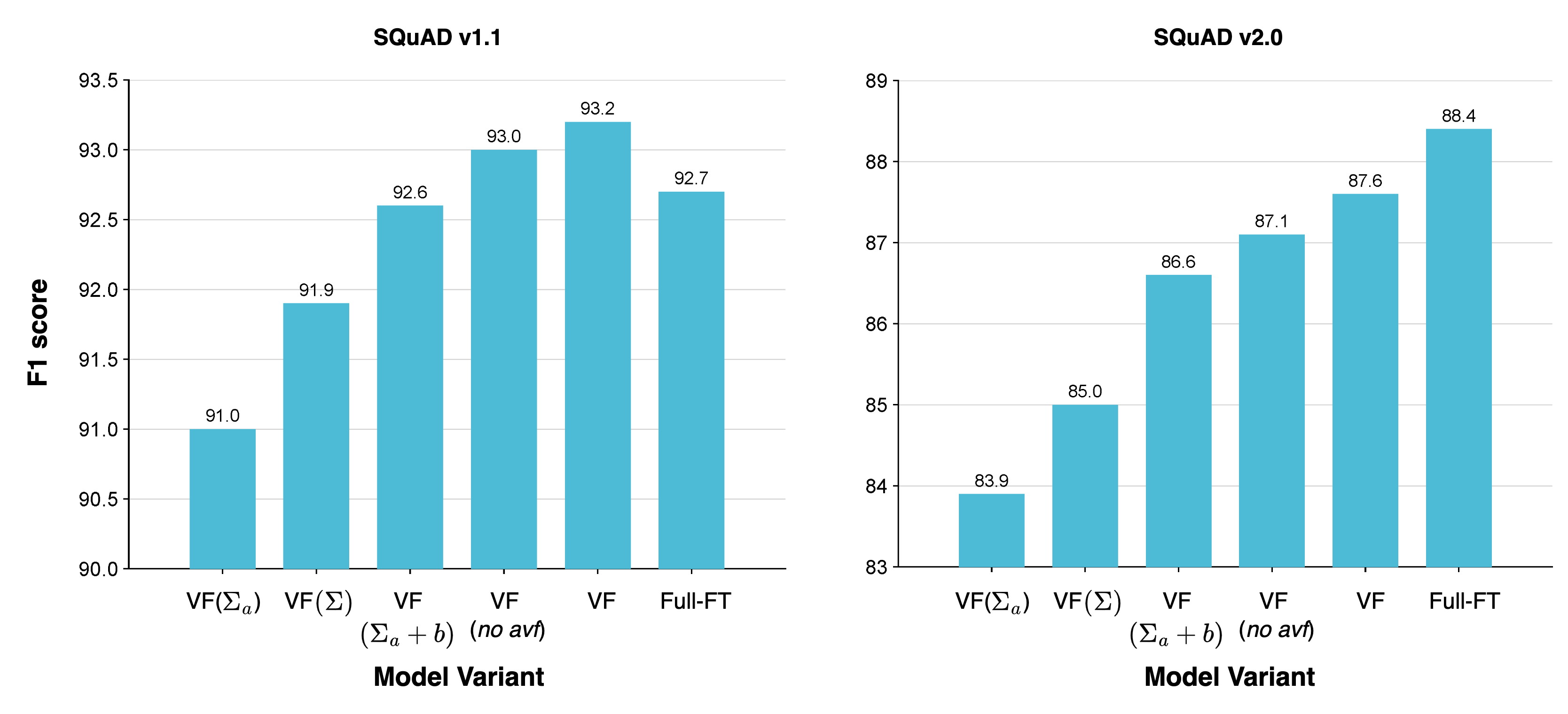}}
\caption{Ablation study about AVF and different trainable vectors configuration. We report the F1 scores for SQuAD v1.1 and SQuAD v2.0 datasets of QA task.}
\label{qa_abl}
\vskip 0.5in
\end{figure}

Figure \ref{qa_abl} presents the results of fine-tuning the DeBERTaV3-base model across five variants mentioned above on QA tasks. On the SQuADv1.1 dataset, the performance difference between VectorFit (\textit{no AVF}) and VectorFit is 0.2\%. On the more challenging SQuADv2.0 dataset, this difference increases to 0.5\%, highlighting the critical role of AVF. Additionally, the average 1\% performance gap between VectorFit ($\Sigma_a$) and VectorFit ($\Sigma$) underscores the importance of singular vectors associated with the fully connected modules ($f_1$ and $f_2$) in the transformer block. Lastly, the performance difference between VectorFit ($\Sigma_a$) and VectorFit ($\Sigma_a + b$) emphasizes the significance of training the bias vectors.

Figure \ref{glue_abl} provides a similar analysis on the GLUE benchmark, yielding results consistent with the observations discussed earlier. Further examination of the training strength of various vectors for different VectorFit variants fine-tuned on the COLA dataset is included in Appendix \ref{app_mexp_ts_abl}.

\subsection{Efficacy of AVF Against Other Possible Approaches}

In this section, we present the experiments conducted by replacing AVF with other possible approaches. We explore L1 regularization and Random vector freezing, where we randomly freeze trainable vectors. From Table \ref{l1_avf} it can be seen that AVF consistently outperforms the other two possible approaches. A key distinction between AVF and L1 regularization lies in the granularity at which they operate. Standard L1 regularization regularizes individual singular values, whereas AVF targets the overall strength of training of singular vectors and bias vectors. We hypothesize that regularizing individual singular values may constrain the model’s expressiveness, as it indirectly limits the flexibility of the associated left and right singular vectors. Random vector freezing performs better than L1 regularization but it leads to slower and unstable training. On the other hand, AVF leads to higher performance with stable training.

\subsection{Limitations}
\label{sec_dis_lim}
Although VectorFit demonstrates exceptional performance with high parameter efficiency, the AVF mechanism's effectiveness depends on careful hyperparameter selection, a challenge shared by comparable methods like AdaLoRA. To overcome this, we provide some heuristics for hyperparameter selection in Appendix \ref{app_impl}. Another limitation is that the number of trainable parameters is currently bounded, as no new parameterized weights are introduced. In future work, we aim to address this by exploring the parameterization of left and right singular matrices, potentially increasing the upper limit of trainable parameters and further enhancing the method's flexibility and performance.

\section{Conclusion}
\label{sec_conc}
We introduce VectorFit, a novel PEFT approach that extracts meaningful singular vectors from weight matrices using SVD and adaptively trains the singular and bias vectors. This method enables high-rank and intrinsic knowledge-aware adaptation of pre-trained models, significantly enhancing both model performance and parameter efficiency. Through comprehensive experiments across diverse language and vision tasks, we demonstrate that VectorFit surpasses existing methods in terms of performance as a function of parameter efficiency. Also, utilizing VectorFit to fine-tune PFMs for downstream tasks is straightforward and cost effective. Additionally, we provide extensive theoretical and empirical insights into its operation to enable further research in this area. In future, we plan to conduct mathematical analysis of weight matrix transformations during fine-tuning, aiming to develop novel parameterization strategies beyond singular vectors and biases.





{\footnotesize
\bibliography{mybibfile}}


\newpage
\appendix
\onecolumn
\begin{huge}\textbf{Appendix}\end{huge}
\section{Memory Usage}
\label{app_A}
\begin{figure}[!h]
\centerline{\includegraphics[scale=0.5]{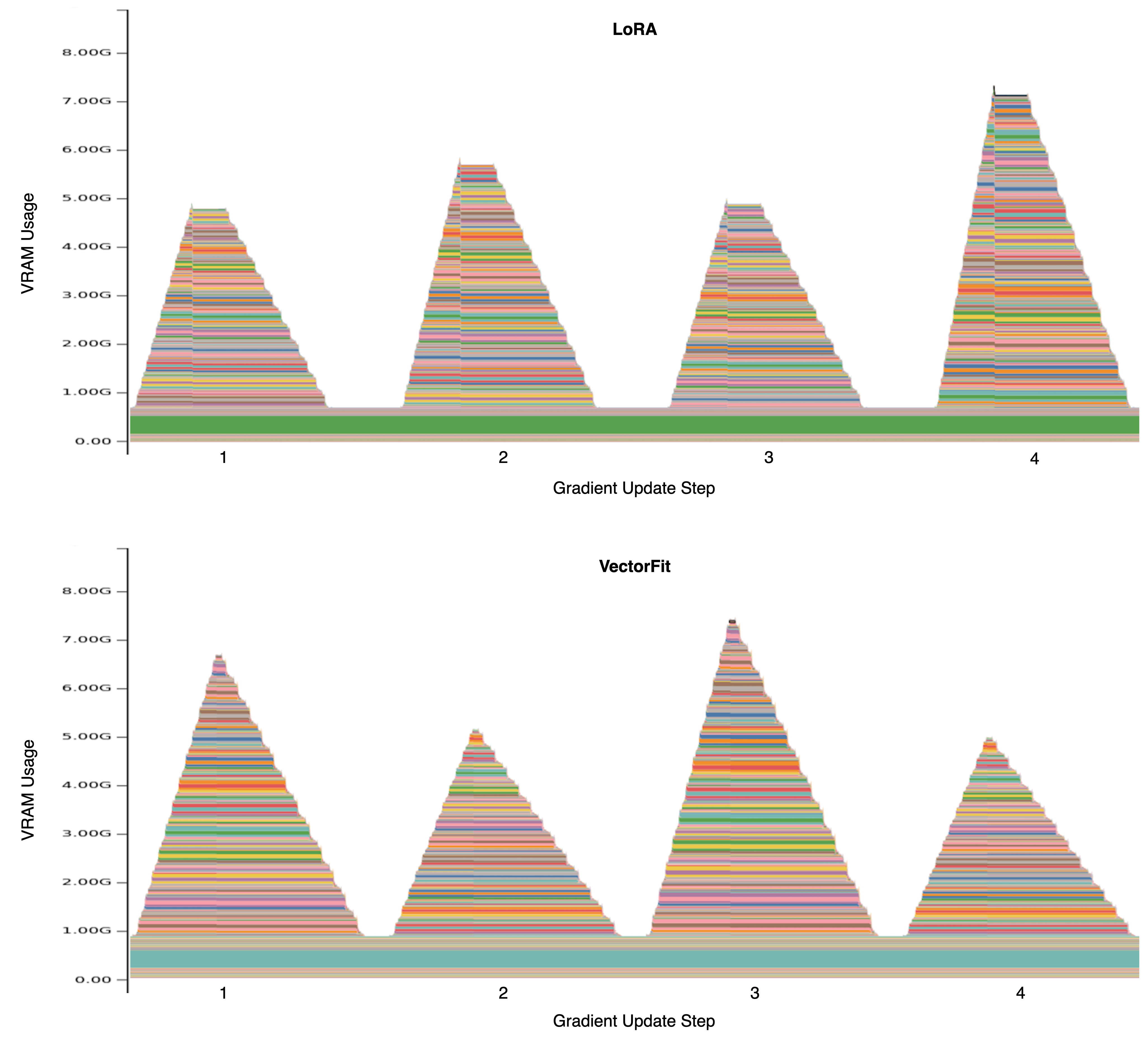}}
\caption{PyTorch memory trace \cite{memviz} comparison of 4 training steps for LoRA (r = 1) on the top and VectorFit on the bottom.}
\label{memtrace}
\end{figure}

\begin{figure}[!h]
\vskip 0.2in
\centerline{\includegraphics[scale=0.7]{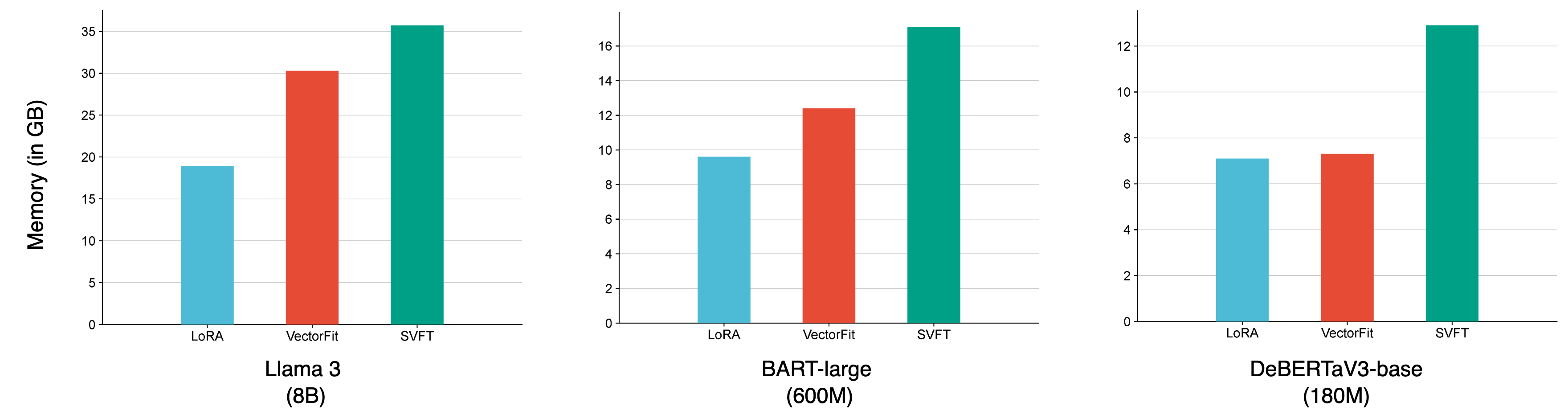}}
\caption{Memory usage.}
\label{memuse}
\end{figure}

As discussed in Section \ref{sec_vft}, while retaining the left and right singular matrices increases the overall parameter count, the practical impact is minimal. Figure \ref{memtrace} compares the GPU memory usage of VectorFit and LoRA (r = 1) on the MNLI dataset using DeBERTaV3-base with full precision training. The figure demonstrates that both methods use comparable amounts of memory, where VectorFit requires approximately 200MB of additional memory using 0.08\% of trainable parameters. Figure \ref{memuse} shows the GPU memory consumption comparison between LoRA, VectorFit, and SVFT with different base models. We observe that VectorFit consumes higher memory compared to LoRA on larger models while using significantly less memory than that of SVFT. The experiments were conducted on an Nvidia A100 GPU with 40GB of RAM.

\section{Training Speed}
\label{app_B}
We fine-tune DeBERTaV3-base on the MNLI dataset with full precision and on the SQuAD v2.0 dataset with mixed precision training to assess the training speed of VectorFit, measured as the time required to train one epoch. Table \ref{speed_comp} shows that VectorFit reduces training time by 17.5\% on MNLI and 16.6\% on SQuAD v2.0 compared to baseline. This improvement is due to VectorFit's simpler computational graph compared to other methods, resulting in faster processing. This experiment was conducted on an Nvidia Titan XP GPU with 12GB of RAM.

\begin{table*}[!h]
\caption{Comparison of practical training time.}
\label{speed_comp}
\vspace{0.5cm}
\centering

\begin{tabular}{c|c|c|c}
\toprule
Dataset                     & \# Params & Method             & Time / Epoch \\ \midrule
\multirow{5}{*}{MNLI}       & 0.08\%    & LoRA               & 82 min       \\
                            & 0.08\%    & AdaLoRA            & 91 min       \\
                            & 0.08\%    & VectorFit          & 75 min       \\
                            & 0.07\%    & VectorFit ($\Sigma_a + b$) & 71 min       \\
                            & 0.01\%    & VectorFit ($\Sigma_a$) & 64 min       \\ \midrule
\multirow{5}{*}{SQuAD v2.0} & 0.08\%    & LoRA               & 98 min       \\
                            & 0.08\%    & AdaLoRA            & 108 min      \\
                            & 0.08\%    & VectorFit          & 90 min       \\
                            & 0.07\%    & VectorFit ($\Sigma_a + b$) & 84 min       \\
                            & 0.01\%    & VectorFit ($\Sigma_a$) & 75 min       \\ \bottomrule
\end{tabular}
\end{table*}

\section{Implementation Details}
\label{app_impl}

This section outlines the implementation details of our method and the baselines used in various experiments. Most of our experiments were conducted on the NVIDIA A100(40G) GPU. We employ the AdamW optimizer with \(\beta_1 = 0.9\) and \(\beta_2 = 0.999\), no warmup, and no weight decay for all our experiments. For the AVF-related hyperparameters, we adopt the following values as a general guideline:
\begin{itemize}
    \item \(t_i\): Approximately 11 epochs' worth of training steps to ensure proper warm-up for all trainable vectors.
    \item \(t_f\): Approximately 1 epoch's worth of training steps to allow for significant updates to the trainable vectors.
    \item \(k \leq 5\): As this value is generally observed to yield the best performance with stable training.
\end{itemize} 

\subsection{Natural Language Understanding}
\label{app_impl_NLU}
Table \ref{hp_glue} gives the hyperparameters used for each task in GLUE benchmark. We experimented using the following learning rates ($1e-2, 1e-3, 1e-4, 3e-4, 5e-4$) and observed that $1e-3$ works best for all tasks in GLUE.
\begin{table*}[!h]
\caption{Hyperparameter setup of VectorFit for GLUE benchmark.}
\vspace{0.5cm}
\label{hp_glue}
\centering
\begin{tabular}{c|ccccccc}
\toprule
Dataset & learning rate & epochs & batch size & $t_i$ & $t_f$ & $n_f$ & $k$ \\ \midrule
MNLI & $1e-03$ & 20  & 32 & 135000 & 10000 & 5 & 5 \\
SST2 & $1e-03$ & 30  & 32 & 23200  & 2100  & 10 & 5 \\
COLA & $1e-03$ & 35  & 32 & 3000   & 200   & 5  & 5 \\
QQP  & $1e-03$ & 25  & 32 & 125100 & 11000 & 10 & 5 \\
QNLI & $1e-03$ & 25  & 32 & 36000  & 3200  & 10 & 5 \\
RTE  & $1e-03$ & 50 & 32 & 800    & 70    & 27 & 5 \\
MRPC & $1e-03$ & 50 & 32 & 1260   & 110   & 27 & 5 \\
STSB & $1e-03$ & 50  & 32 & 1900   & 180   & 27 & 5 \\ \bottomrule
\end{tabular}
\end{table*}

Table \ref{hp_glue_bl} presents the hyperparameters related to budget allocation of the baselines. $d$ is the hidden dimension for the adapters, $r$ is the rank of LoRA incremental weight matrices, and $b^{(T)}$ is the target budget of AdaLoRA. We use SVFT with random setting and $d = 2$.

\begin{table*}[!h]
\caption{Budget setup of baselines for GLUE benchmark.}
\label{hp_glue_bl}
\begin{center}
\begin{small}
\begin{tabular}{l|cccc}
\toprule
{\# Params} & {Houlsby Adapter ($d$)} & {Pfeiffer Adapter ($d$)} & {LoRA ($r$)} & {AdaLoRA ($b^{(T)}$)}
\\
\midrule
{1.2M} & {32} & 64 & 8 & 576 
\\
{0.6M} & 16 & 32 & 4 & 288 
\\
{0.3M} & 8 & 16 & 2 & 144 
\\
\bottomrule
\end{tabular}
\end{small}
\end{center}
\end{table*}

\subsection{Question Answering}
\label{app_impl_QA}
Table \ref{hp_qa} gives the hyperparameters used for each dataset of QA task. We experimented using the following learning rates ($1e-2, 1e-3, 1e-4, 3e-4, 5e-4$) and observed that $1e-3$ works best for both datasets of QA task. 

\begin{table*}[!h]
\caption{Hyperparameter setup of VectorFit for question answering tasks.}
\vspace{0.5cm}
\label{hp_qa}
\centering
\begin{tabular}{c|ccccccc}
\toprule
Dataset & learning rate & epochs & batch size & $t_i$ & $t_f$ & $n_f$ & $k$ \\ \midrule
Squad v1.1       & $1e-03$                  & 20              & 16                  & 60700         & 5500          & 6            & 5          \\
Squad v2.0       & $1e-03$                  & 20              & 16                  & 90300         & 8200          & 6            & 5          \\ \bottomrule
\end{tabular}
\end{table*}

\newpage

Table \ref{hp_qa_bl} presents the hyperparameters related to budget allocation of the baselines.

\begin{table*}[!h]
\caption{Budget setup of baselines for QA tasks.}
\label{hp_qa_bl}
\begin{center}
\begin{small}
\begin{tabular}{c|ccccc}
\toprule
{\# Params} & {Houlsby Adapter $(d)$} & {Pfeiffer Adapter $(d)$} & {LoRA $(r)$} & {AdaLoRA $(b^{(T)})$} & {SVFT $(d)$}
\\
\midrule
{0.08\%} & 4 & 8 & 1 & 72 & 1
\\
\bottomrule
\end{tabular}
\end{small}
\end{center}
\end{table*}

\subsection{Natural Language Generation}
\label{app_impl_NLG}
Table \ref{hp_nlg} gives the hyperparameters used for each dataset of NLG task. We experimented using the following learning rates ($1e-2, 1e-3, 1e-4, 3e-4, 5e-4$) and observed that $1e-3$ works best for both datasets of NLG task.
\begin{table*}[!h]
\caption{Hyperparameter setup of VectorFit for natural language generation tasks.}
\vspace{0.5cm}
\label{hp_nlg}
\centering
\begin{tabular}{c|ccccccc}
\toprule
Dataset & learning rate & epochs & batch size & $t_i$ & $t_f$ & $n_f$ & $k$ \\ \midrule
XSum             & $1e-03$                  & 30              & 64                  & 35070         & 3100          & 10            & 5          \\
CNN/Dailymail    & $1e-03$                  & 30              & 64                  & 31500         & 4400          & 10            & 5          \\ \bottomrule
\end{tabular}
\end{table*}

Table \ref{hp_nlg_bl} presents the hyperparameters related to budget allocation of the baselines used for experiments with Xsum and CNN/Dailymail datasets for NLG task.

\begin{table*}[!h]
\caption{Budget setup of baselines for NLG tasks.}
\label{hp_nlg_bl}
\begin{center}
\begin{small}
\begin{tabular}{ccccc}
\toprule
{Houlsby Adapter $(d)$} & {Pfeiffer Adapter $(d)$} & {LoRA $(r)$} & {AdaLoRA $(b^{(T)})$} & {SVFT $(d)$}
\\
\midrule
8 & 16 & 2 & 144 & 2
\\
\bottomrule
\end{tabular}
\end{small}
\end{center}
\end{table*}

\subsection{Image Classification}
\label{app_impl_ic}
Table \ref{hp_ic} gives the hyperparameters used for each dataset of image classification task. We experimented using the following learning rates ($1e-2, 1e-3, 1e-4, 3e-4, 5e-4$) and the best performing learning rates are given in the table.
\begin{table}[!h]
\caption{Hyperparameter setup of VectorFit for image classification tasks.}
\vspace{0.4cm}
\label{hp_ic}
\centering
\begin{tabular}{@{}c|ccccccc@{}}
\toprule
Dataset  & learning rate & epochs & batch size & $t_i$ & $t_f$ & $n_f$ & $k$ \\ \midrule
CIFAR10  & $1e-03$         & 20     & 128        & 3600 & 300  & 4    & 5 \\
GTSRB    & $1e-03$         & 20     & 128        & 1900 & 170  & 4    & 5 \\
MNIST    & $1e-02$         & 20     & 128        & 4300 & 350  & 4    & 5 \\
RESISC45 & $1e-02$         & 20     & 128        & 2300 & 200  & 4    & 5 \\ \bottomrule
\end{tabular}
\end{table}

For the baselines, we use LoRA with $r = 2$, AdaLoRA with $b^{(T)} = 144$, and SVFT with $d = 2$.

\subsection{Image Generation}
\label{app_impl_ig}
Dreambooth fine-tuning for various subjects in the dataset were done using prior preservation loss with the weightage varying between 0.5 to 1.0 depending on the subject. We use 300 class images for each subject, a learning rate of $5e-5$, and a batch size of 4. We use the rank of 2 for fine-tuning with LoRA.

\section{Additional Experiments}
\label{app_mexp}
\subsection{Training Strength Ablation}
\label{app_mexp_ts_abl}
\begin{figure*}[!h]
\begin{center}
\centerline{\includegraphics[scale=0.5]{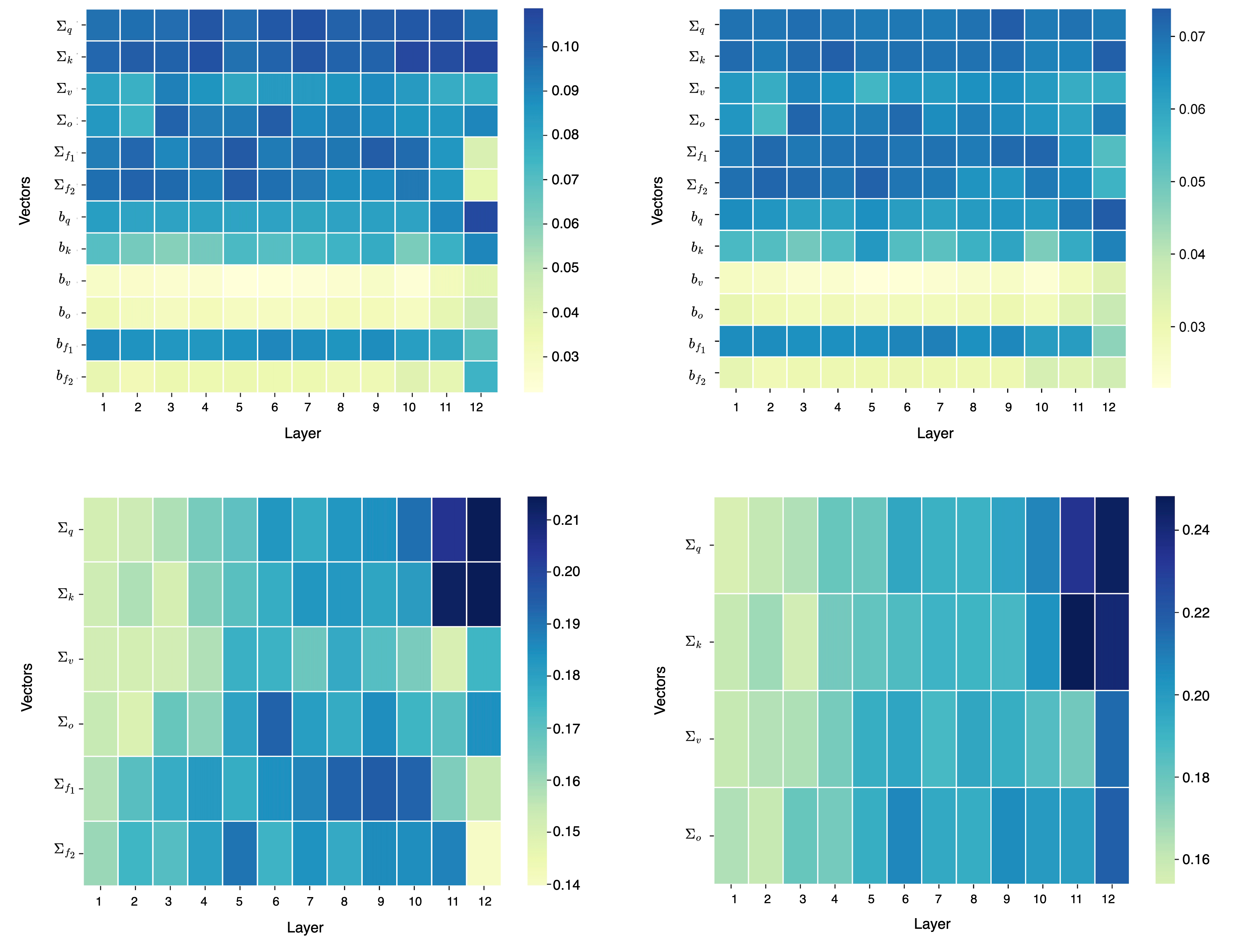}}
\caption{The training strength $S_v$ of each trainable vector after fine-tuning of DeBERTaV3-base on the COLA dataset is shown for VectorFit without AVF (top-left), VectorFit with AVF (top-right), VectorFit ($\Sigma$) (bottom-left), and VectorFit ($\Sigma_a$) (bottom-right). The x-axis represents the layer index, while the y-axis corresponds to different types of trainable vectors.}
\label{ts_abl}
\end{center}
\vskip -0.2in
\end{figure*}

Figure \ref{ts_abl} shows the training strength heatmap of various trainable vectors for different variants of VectorFit. We can observe that VectorFit with AVF (top-right) achieves the most equitable training possible among the trainable vectors and hence maintains an overall lower training strength. We can also observe that as the number of trainable vectors is reduced, the training strength of the vectors increases to make up for the reduced number of trainable parameters.

\subsection{NLU Tasks Ablation}
\label{app_mexp_qa_abl}

\begin{figure*}[!h]
\begin{center}
\centerline{\includegraphics[width=\textwidth]{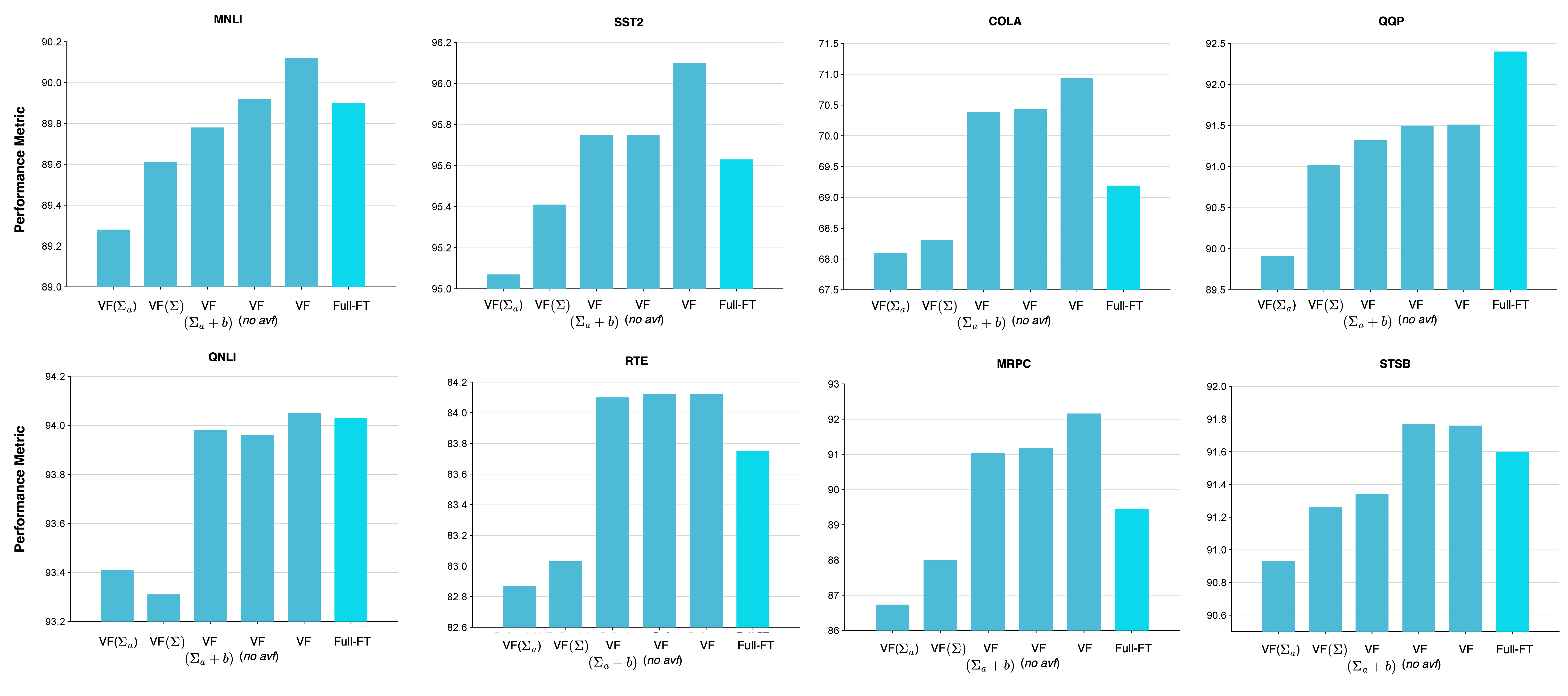}}
\caption{Ablation study about AVF and different trainable vectors configuration on the GLUE benchmark. We report the matched accuracy for MNLI, Matthew’s correlation for CoLA, Pearson correlation for STS-B, and accuracy for the other tasks.}
\label{glue_abl}
\end{center}
\vskip -0.2in
\end{figure*}

Figure \ref{glue_abl} shows the ablation graphs for the GLUE benchmark with all five variants of our method. The graphs show the efficacy of AVF where VectorFit with AVF gives a higher performance on all the datasets. 

\newpage

\subsection{QA Tasks Ablation}
\label{app_mexp_qa_abl}

Table \ref{abltn-qa-table} presents the performance of various VectorFit variants. Notably, the most parameter-efficient version, VectorFit(\(\Sigma_a\)), which uses only 0.01\% of trainable parameters, achieves up to 98\% of the F1 score obtained with Full-FT.

\begin{table}[!ht]
\caption{Ablation study on QA.}
\vspace{0.5cm}
\label{abltn-qa-table}
\centering
\begin{tabular}{c|c|cc}
\toprule
\textbf{Model} & \textbf{\# Params} & \textbf{Squad v1.1 (EM/F1)} & \textbf{Squad v2.0 (EM/F1)} \\ \midrule
VectorFit ($\Sigma_a$)       & 0.01\% & 83.8/91.0 & 80.2/83.9 \\
VectorFit ($\Sigma$) & 0.02\% & 84.9/91.9 & 81.6/85.0 \\
VectorFit ($\Sigma_a + b$)   & 0.07\% & 86.4/92.6 & 83.7/86.6 \\
VectorFit (\textit{no avf})  & 0.08\% & 86.7/93.0 & 84.2/87.1 \\
VectorFit           & 0.08\% & 87.0/93.2 & 84.4/87.6 \\ \bottomrule
\end{tabular}
\end{table}

\newpage

\subsection{Rank Analysis Continued}
\label{app_mexp_rank}

\begin{figure*}[!h]
\centerline{\includegraphics[scale=0.23]{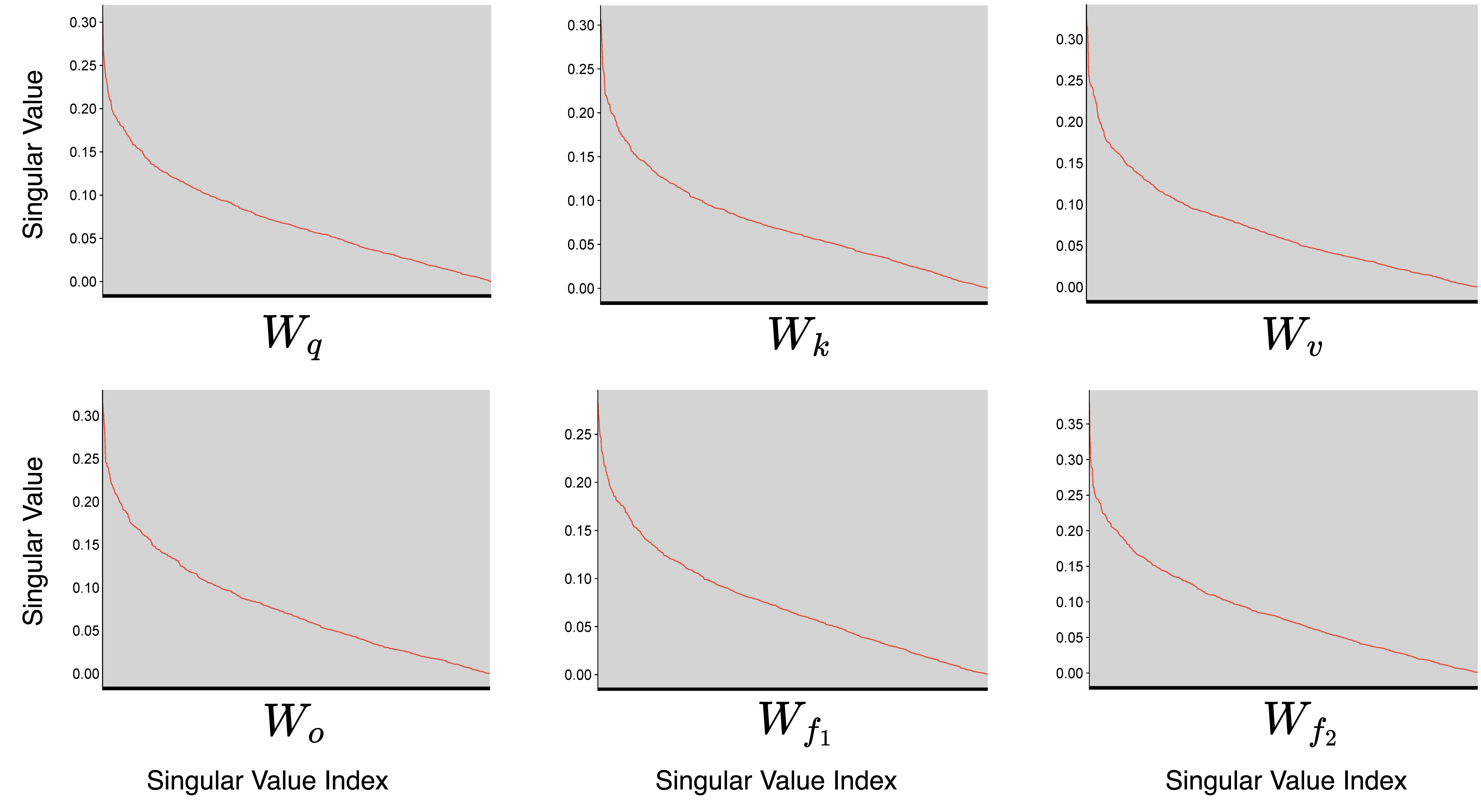}}
\caption{Singular value graphs of $\Delta^*$ for all the modules of a randomly picked layer (layer 6) of ViT-base model fine-tuned with VectorFit on CIFAR10 dataset. X-axis represents the singular value position/index and Y-axis represents the singular value.}
\label{sin_values_vit}
\vskip -0.2in
\end{figure*}

\begin{figure*}[!h]
\centerline{\includegraphics[scale=0.25]{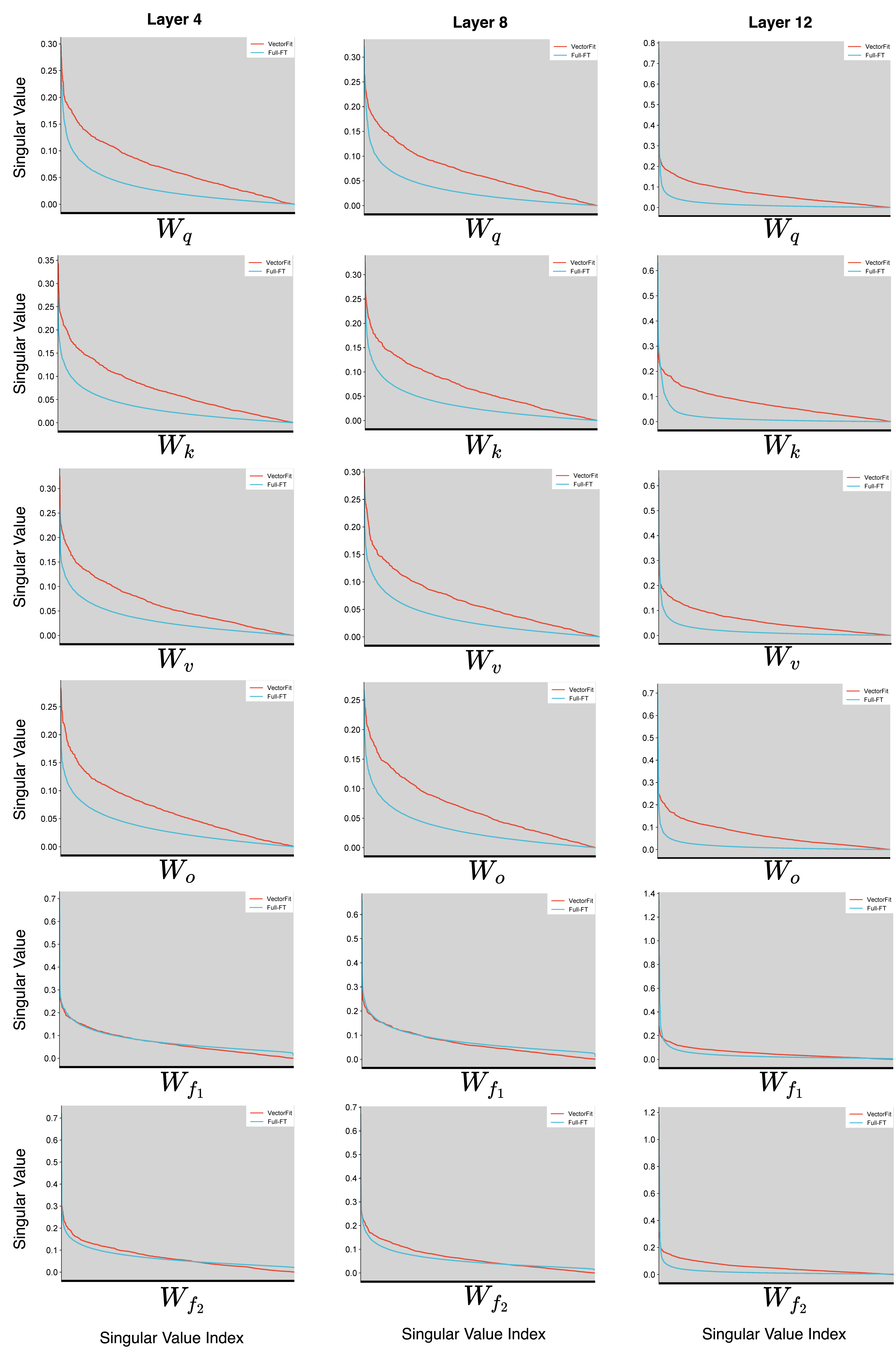}}
\caption{Singular value graphs of $\Delta^*$ for all the modules of randomly picked layers in case of VectorFit and Full-FT. X-axis represents the singular value position/index and Y-axis represents the singular value.}
\label{sin_values}
\vskip 0.2in
\end{figure*}

Figure \ref{sin_values} presents the singular value distributions of the \(\Delta^*\) matrices discussed in Section \ref{sec_dis_ra}. For the DeBERTaV3-base model, each singular vector is 768-dimensional, and all 768 singular values are plotted. The graphs reveal that \(\Delta^*\) for Full-FT is not inherently low-rank, as even the smallest singular values remain non-zero for many weight matrices. Additionally, the plots demonstrate that VectorFit achieves high-rank adaptation, closely approximating Full-FT for several weight matrices. 
Figure \ref{sin_values_vit} shows the singular values of \(\Delta^*\) of ViT-base model fine-tuned with VectorFit on CIFAR10 dataset.

\newpage

\subsection{Weight Transformation During VectorFit Fine-Tuning}
\label{app_wtransformation}

\begin{figure*}[!h]
\centerline{\includegraphics[width=\textwidth]{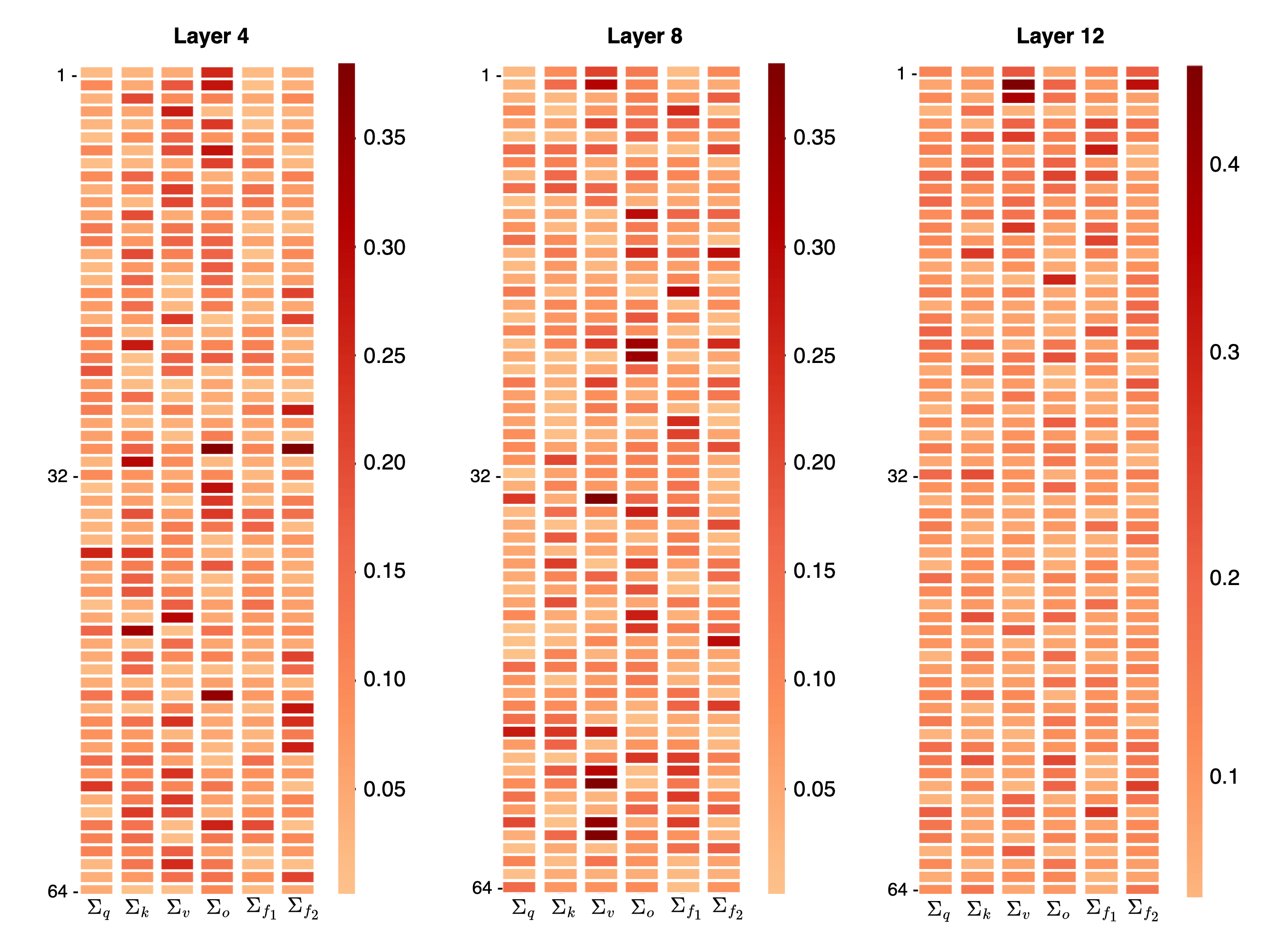}}
\caption{Heatmap representing the variations in first 64 singular values of different singular vectors before and after fine-tuning. The heatmaps are generated with randomly picked layers of DeBERTaV3-base model fine-tuned using VectorFit on COLA dataset.}
\label{sin_values_hm}
\vskip -0.2in
\end{figure*}

The heatmap of variation of the first 64 singular values before and after full fine-tuning in each singular vector of randomly selected layers is displayed in Figure \ref{sin_values_hm}. This depicts the weight matrix's stretching in its multi-dimensional hyper-space. It should be noted that upon fine-tuning, 
even the least significant singular directions might become the principal singular directions. This shows that VectorFit is highly expressive.

\begin{figure*}[!h]
\centerline{\includegraphics[scale=0.8]{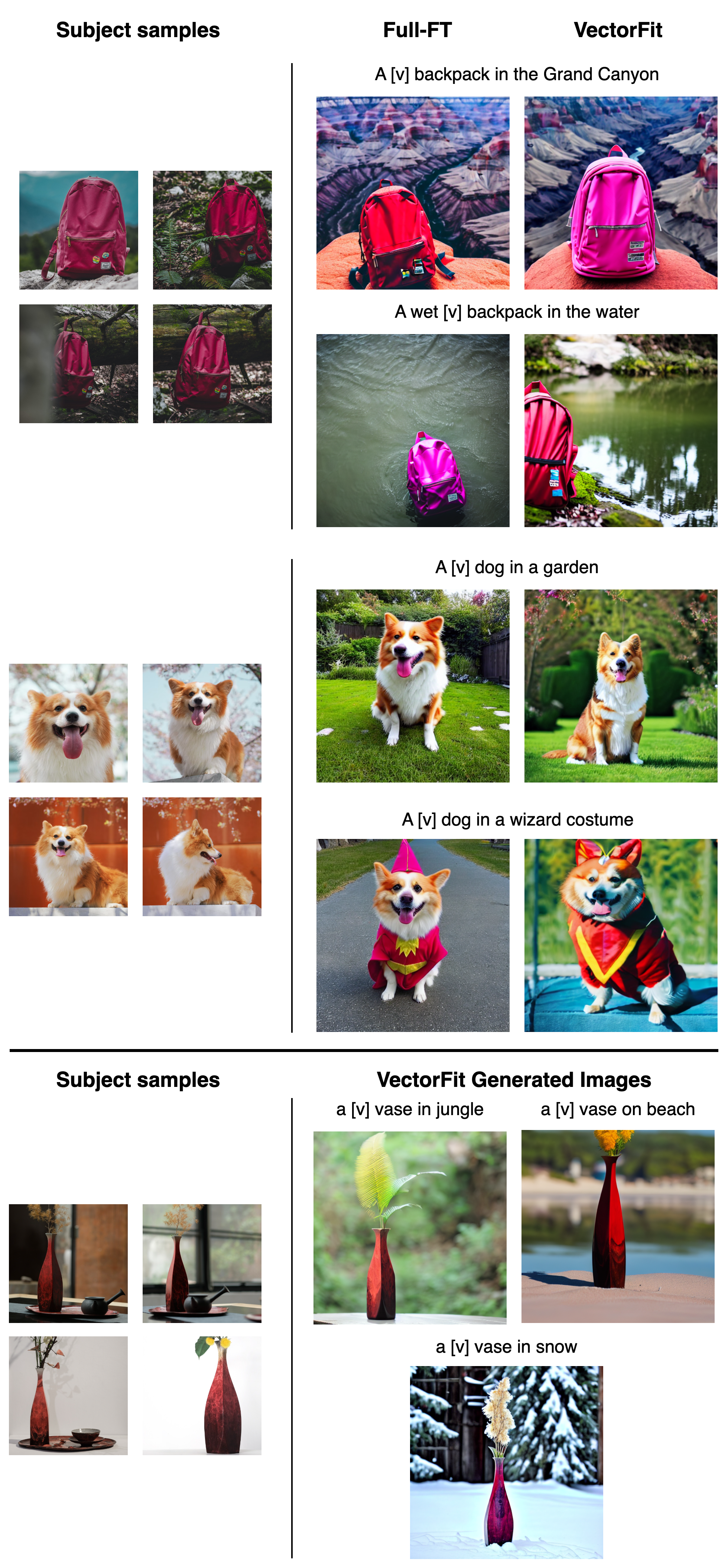}}
\caption{Visual comparison of images generated by Stable Diffusion v1.4 fine-tuned with VectorFit and Full-FT methods using Dreambooth approach for Subject-driven Image generation.}
\label{db_qual}
\vskip -0.2in
\end{figure*}



\end{document}